\documentclass{IEEEtran}
\usepackage{times}
\usepackage{epsfig}
\usepackage{graphicx}
\usepackage{amsmath}
\usepackage{amssymb}
\usepackage{float}
\usepackage{cite}
\usepackage{url}
\usepackage[utf8]{inputenc}
\UseRawInputEncoding
\usepackage{authblk}

\ifCLASSINFOpdf
\else
\fi

\begin{document}

\title{Semantic Segmentation of Periocular Near-Infra-Red Eye Images Under Alcohol Effects}

\author[1,6]{Juan Tapia}
\author[2]{Enrique Lopez Droguett}
\author[3,4]{Andres Valenzuela, Daniel Benalcazar}
\author[5]{Leonardo Causa}
\author[6]{Christoph Busch}
\affil[1]{Universidad de Chile and Hochschule Darmstadt - Germany,\authorcr Email: {\tt jtapiafarias@ing.uchile.cl}}
\affil[2]{Department of Civil and Environmental Engineering, and Garrick Institute for the Risk Sciences, University of California, Los Angeles, USA.}
\affil[3,4]{Universidad de Chile, DIMEC, Chile,}
\affil[5]{TOC Biometrics, R+D Center, Chile,}
\affil[6]{da/sec-Biometrics and Internet Security Research Group, Hochschule Darmstadt, Germany.        

\\\textbf{This work has been submitted to the IEEE for possible publication. Copyright may be transferred without notice, after which this version may no longer be accessible}} 




\maketitle

\begin{abstract}
This paper proposes a new framework to detect, segment, and estimate the localization of the eyes from a periocular Near-Infra-Red iris image under alcohol consumption. The purpose of the system is to measure the fitness for duty. Fitness systems allow us to determine whether a person is physically or psychologically able to perform their tasks. Our framework is based on an object detector trained from scratch to detect both eyes from a single image. Then, two efficient networks were used for semantic segmentation; a Criss-Cross attention network and DenseNet10, with only 122,514 and 210,732 parameters, respectively. These networks can find the pupil, iris, and sclera. In the end, the binary output eye mask is used for pupil and iris diameter estimation with high precision. Five state-of-the-art algorithms were used for this purpose. A mixed proposal reached the best results. A second contribution is establishing an alcohol behavior curve to detect the alcohol presence utilizing a stream of images captured from an iris instance. Also, a manually labeled database with more than 20k images was created. Our best method obtains a mean Intersection-over-Union of 94.54\% with DenseNet10 with only 210,732 parameters and an error of only 1-pixel on average.
\end{abstract}


\IEEEpeerreviewmaketitle

\section{Introduction}
\IEEEPARstart{T}{he} concept of biometrics is the set of intrinsical and behavioral characteristics that can be used to verify the identity of an individual. All human beings have unique morphological characteristics that differentiate and identify us, the shape of the face, the geometry of parts of our body, like hands, our eyes, and perhaps the best known, the fingerprint \cite{anil,Daugman2004irisrecognition}. 

Biometric features extracted from images representating such morphological characterisitcs. For example, the face is a variable characteristic that depends on many factors, which can be easily changed by involuntary or voluntary action of the individual, simulating sad/happy faces or others. Fingerprint recognition has reached a significant deployment, but its vulnerability in identity theft has been demonstrated even more so with people who made handcrafted labor actively to carry out their tasks due to the fingerprint's deterioration.

Iris recognition systems have been used mainly to recognize the cooperative subjects in controlled environments for borderline, demographics and to gain access to building and office using near-infra-red capture devices \cite{Daugman2004irisrecognition, sommaly,Tapia2016GenderClassification}. With the improvements in iris performance and reduction in the cost of iris acquisition devices, the technology will witness broader applications and may be confronted with newer challenges. One kind of this new challenge is identifying if a capture subject is under alcohol, drug effects, or even in conditions of sleep deprivation and sleep restriction. This area is known as "Fitness for Duty" \cite{FFD,murphy} and allows to determine whether the person is physically or psychologically able to perform their task. \cite{murphy}. This system delivers an answer based on the statistical people behavior threshold of "fit" and "unfit". It does not have a relationship with the automatic alcohol test that measures alcohol percentage on blood.

To determine whether a worker is fit to perform a task, it is necessary to observe that repetitive behaviors or biometric factors are manifested through their body that allows establishing the relationship between the cause and the effect in his/her behavior. The iris and pupil movements are controlled by the Central Nervous Systems (CNS)\cite{Adler1985}. In this condition, the subject can not change as they wish the pupil or iris movement. This action is initiated automatically for an external factor such as light or an internal factor, such as alcohol consumption and others. Therefore, the iris is highly reliable to measure the fitness for duty of the subject. This technique has been previously used but with optical measurements using a light beam to measure the changes of the iris, similar to the driver's license test \cite{SunYaoJi2012}. 

In order to create a framework to assess the worker’s fitness for duty based on iris measurements is necessary to capture $N$ frames, detect both eyes, and segment the images to localize the pupil and the iris to measure the changes over time. To segment this kind of image is not a trivial task because the method needs to be efficient in the number of parameters in order to be implemented in a regular iris sensor. Most of these sensors are mobile devices self-integrated with a limited size of memory.

State-of-the-art algorithms such as semantic segmentation have been mainly trained to localize very complex objects from cities such as cars, buildings, and people and a few in biometric gaze applications. Even when we find more than one class object in the same image. Traditional pre-trained implementation models reach meager results when directly applied to eye segmentation and gaze estimation without alcohol presence. Another limitation is that most state of the art segmentation algorithms are based on deep convolutional networks with a large number of layers and parameters \cite{Badrinarayanan2017SegNet, Chen2018AtrousSeparableConvolution}.

Recently, automatic pupil segmentation is attracting many researchers to find the precise measurement of the pupil radii to apply to biometric analytics as part of medical, entertainment applications, virtual reality lens, among others \cite{sclera_comp}. In the biomedical field, there is a vital requirement in developing precise and automatic segmentation systems to capture saccade velocity, latency, the diameter of the iris, and pupil \cite{Czajka2015pupildynamics, Pinheiro2015alcoholvideo, arora2012alcohol}.

While alcohol influence on iris recognition has not been extensively investigated, not much work has been done to analyze and counteracting the effect of drug-induced pupil dilation and sleepiness on iris images \cite{arora2012alcohol}. 
This work focuses on pupil and iris behavior induced by the use of alcohol agents, but it could also be extended and allied to drug consumption as it has been shown that drug induced pupil dilation affects iris recognition performance negatively. However, no solution has been investigated to counteract its effect on iris recognition. Note that alcohol may be used by an adversary to mask their identity from an iris recognition system \cite{arora2012alcohol}. These agents can be easily obtained online without a medical prescription. Hence, there is a need to understand and counteract the effect of alcohol on iris recognition.

There are many issues associated with alcohol consumption and sleep deprivation when it comes to daily consumption: reduced performance worker, fatigue driving a car, problem on social and professional situations \cite{RoehrsandRoth}. Therefore, there is always a risk of driving accidents and cause injuries or death due to alcohol overconsumption \cite{Akerstedt2000}. Also, it is needed to account for problems due to the long-term exposure to alcohol like neurotoxicity, heart, liver, immune system, and other organs \cite{Pinheiro2015alcoholvideo}.

An estimated 15-25\% of the workforce works in shifts \cite{BALASUBRAMANIAN202052, WICKWIRE20171156, ijerph16173081}. Working in rotating shifts with work at night can be a significant risk factor. Also, several studies show an essential correlation between shiftwork, work at night, and alcohol consumption as such important triggers for occupational accidents \cite{Richter2021}. Workplace alcohol use and impairment directly affect an estimated 15\% of the U.S. workforce; about 10.9\% work under the influence of alcohol or with a hangover \cite{Frone}. The Australian Government alcohol guidelines report shows 13\% of shift-workers, and 10\% of those on standard schedules reported consuming alcohol at levels risky for short-term harm \cite{DorrianandSkinner}.

In this context, pupil measurement has been used to assess several cognitive functions, including fatigue, depression, and others. \cite{ROMANOBERGSTROM201481}. 

The iris alcohol segmentation present the following challenges:

\begin{itemize}
    \item The people react differently with the same quantity of alcohol; some present pupil dilation, and other pupils constrict.
    
    \item The average constriction size of the pupil in alcohol presence is over the normal ranges. These changes do not allow used parametrical segmentation methods such as Osiris or commercial software.
    
    \item Most of the time, people present semi-closed eyes. This feature is an extra difficulty.
    
    \item In the presence of alcohol, the volunteer in front of the sensor shows an involuntary disbalance. This adds blurring to the capture images.
 
\end{itemize} 

\begin{figure*}[]
\begin{centering}
\includegraphics[scale=0.4]{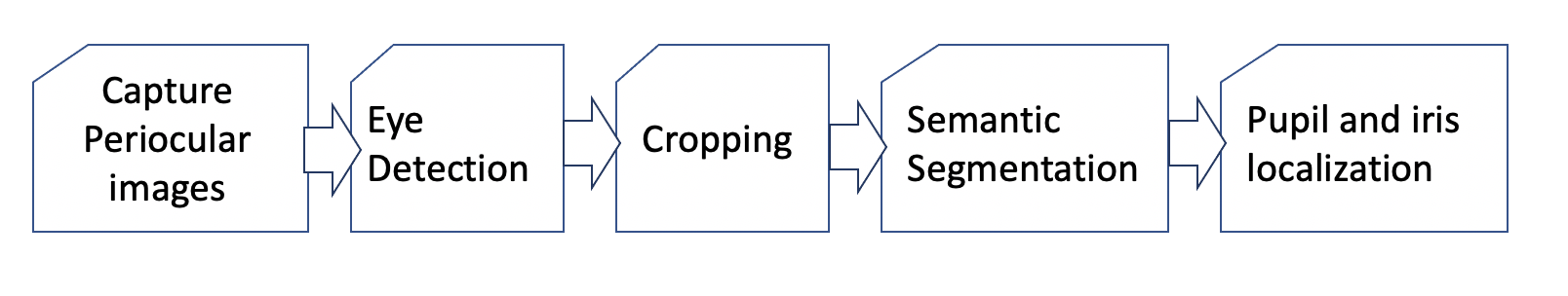}
\par\end{centering}
\caption{\label{farmework1}Block diagram of the proposed framework.}
\end{figure*}

This paper aims to develop an efficient framework to segment and locate the iris and pupil in multiples frames in subjects with and without alcohol consumption. This work is ongoing research whose main objective is to estimate and extract the most relevant features from the iris and pupil to predict the fitness for each person's duties to save lives. See Figure \ref{farmework1}.

\section{Related work}
\label{relate_work}
\subsection{Alcohol}

The influence of alcohol, in particular in iris recognition, was reported by Arora et al., \cite{arora2012alcohol}. They presented a preliminary study of the impact of alcohol on an Iris recognition system. The experiments were performed on the 'Iris Under Alcohol Influence' database. Results show that when comparing pre and post-alcohol consumption images, the overlap between mated and non-mated compare distance score distributions increases by approximately 20\%. These results were obtained using a relatively small database (220 pre-alcohol and 220 post-alcohol images obtained from 55 subjects). The subjects consumed about 200 ml of alcohol (with a 42\% concentration level) in approximately 15 minutes, and the images were captured 15-20 minutes after alcohol consumption.
This work suggests that about one in five subjects under the influence of alcohol may be able to evade identification by iris recognition systems.

Czajka \textit{et al.}\cite{Czajka2015pupildynamics} applied Hough transform operating on directional image (estimation of an image gradient delivering both a gradient value an its direction). After modeling the pupil size, It was shown as function of number of frames, representing the dynamic of pupil. In this way the pupil dynamic is modeled extracting liveness features.  

Kumar \textit{et al.}\cite{Kumar2010DelhiDB} used a series of steps for preprocessing eye images based on replacing pixels over thresholding, median, Gaussian filtering, and a Canny edge detector. Using these steps, they performed segmentation over the Casia v3 and IITD datasets. This dataset does not present alcohol examples. 

Bernstein \textit{et al.} \cite{BernsteinMendezSunEtAl2017} used a spectrogram images of size $224\times224$ from audio wave-forms to identify the presence of alcohol with Convolutional Neural Networks (CNN) and wearable sensors. They used 80 training images (40 positive, 40 negative) and 20 test images (10 positive, 10 negative) and achieved a test accuracy, after adjusting learning rate, number of iterations, and gradient descent algorithm, as well as the time window and coloration of the spectrograms, of 72\% (n=20, 5 trials). 

Koukio \textit{et al.} \cite{Koukiou} proposed the use of thermal images to identify individuals under the influence of alcohol. They have shown that changes in the eye temperature distribution in intoxicated individuals can be detected using thermal imagery \cite{eyetermal}.

\begin{figure*}[]
\centering
\includegraphics[scale=0.43]{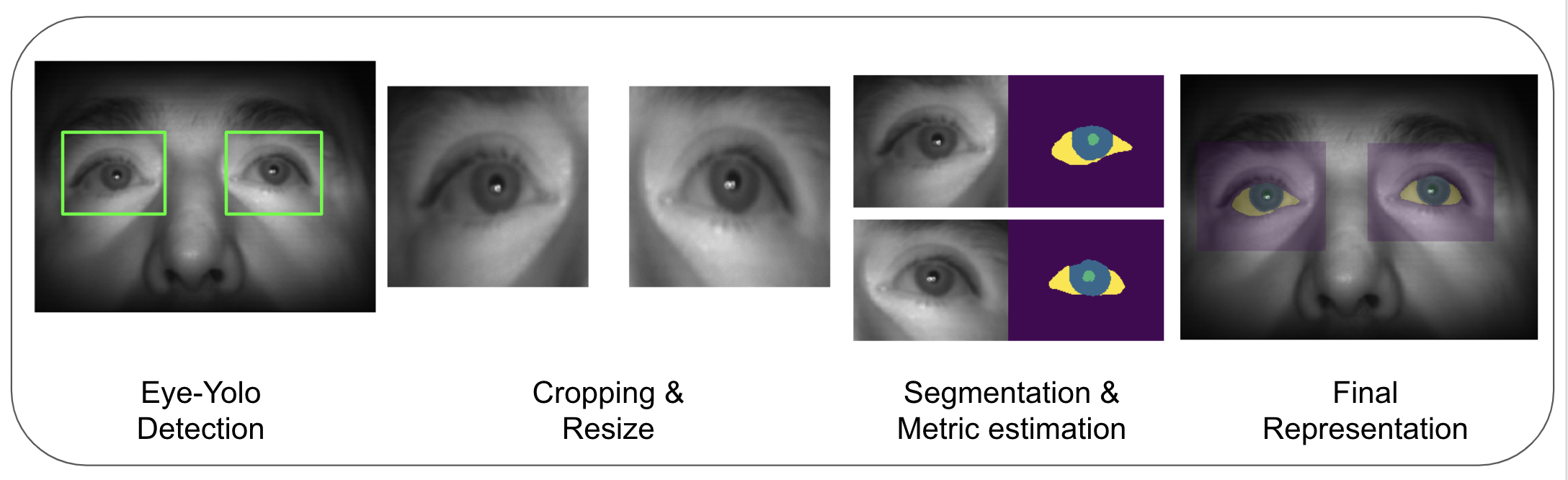}
\caption{\label{all_stages} Graphics demonstration of all the stages used in the proposed framework method based on DenseNet10 and Eye-yolo.}
\end{figure*}

\subsection{Semantic Segmentation}

Semantic segmentation is the process of identify each part/subject in an image. This process is performed pixel by pixel in such a way it is owned by the object contained in the image\cite{valenzuela2020segmentation}. 

The U-Net \cite{Ronneberger2015unet} is a convolutional network architecture for fast and precise segmentation of images to supplement a usual network by successive layers, where upsampling operators replace pooling operators. Hence, these layers increase the resolution of the output. In order to localize, high-resolution features from the path are combined with the upsampled result. A successive convolution layer can then learn to assemble a more precise output based on this information.

Chen \textit{et al.} \cite{Chen2018AtrousSeparableConvolution} proposed a complex DeepLabv3+ as an extension of the previous DeepLabv3 by adding a simple yet effective encoder module to recover the object boundaries.
The rich semantic information is encoded in the output of DeepLabv3 with atrous convolution allowing to control the density of the encoder features depending on a budget of computational resources. Furthermore, the decoder module allows detailed object boundary recovery. DeepLabV3plus+ is a cutting-edge architecture for semantic segmentation. This architecture can do multi-scale processing
without increasing the number of parameters. 
DeepLabV3+ adds an intermediate decoder module on top. After processing the information via DeepLabV3+, the features are then up-sampled $N$ times. 
This network improves the data load from the end of the network and provides a shortcut path from the feature extraction front-end to the near future of the network. However, The DeeplabV3+ uses the ResNet101 as a backbone which is pre-trained using the ImageNet database.

Kumar \textit{et al.} \cite{Badrinarayanan2017SegNet} proposed the first Encoder-Decoder architecture for segmentation tasks called SegNet. SegNet has three main blocks: encoder, hidden vector, and decoder Network, followed by a pixel-wise classification layer. The encoder part will convert the input image into a single-dimensional vector (hidden vector). The decoder network will convert the hidden vector into the corresponding semantic segmentation. This kind of CNN was used to perform sclera segmentation on Multi-Angle Sclera Database (MASD)\cite{Rot2018SegNetOnMASD}. Hassan et al. \cite{hassan2020sipsegnet} used the SegNet to segment pupil, iris, and sclera over several databases.

The Fully Convolutional Neural Networks (FCNN) are the natural evolution of the CNN for segmentation tasks \cite{Long2015FCN, Bezerra2018FCN}. This kind of neural network delivers an image of equal size as the input image containing the segmented classes. To reach that goal, this adds an upsampling layer. Bezerra et al. \cite{Bezerra2018FCN} perform iris segmentation used the following databases: Casia v4, IITD, CrEye-Iris.

The DeepVOG \cite{yiu2019deepvog} is a convolutional layer with $10\times10$ filters which output feature maps with the same size of the input by appropriate padding. The down-sampling path reduces the size of the feature maps and increases the size of receptive fields of convolutional filters each stage, such that more complex features in a larger context can be extracted. 
DeepVOG was developed as a prerequisite for many eye-tracking and Video-OculoGraphy (VOG) methods for accurate pupil localization. DeepVOG is based on Fully Convolutional Neural Network (FCNN). The output simultaneously enables to perform pupil center localization, elliptical contour estimation, and blink detection, all with a single network and with an assigned confidence value, at frame rates above $130 Hz$ on commercial workstations with GPU acceleration. Pupil center coordinates can be estimated with a median accuracy of around 1.0 pixel, and gaze estimation is accurate to within 0.5 degrees. 

Valenzuela \textit{et al.}\cite{valenzuela2020segmentation} proposed an efficient DenseNet based on DenseNet56 and compared several implementation in number of parameters, scores and complexity using DeepLabV3, UNet, Mask-RCNN, DenseNet-56, DenseNet101 and, DenseNet10. These models were trained using the OpenEDS database \footnote{\url{https://research.fb.com/programs/openeds-challenge/}}. The focus was to develop an efficient semantic segmentation method to be implemented in a mobile device. 

Huang \textit{et al.}\cite{ccnet1,ccnet2} proposed a semantic model of full-image dependencies over local feature representations using light-weight computation and memory. They introduce a criss-cross attention module (CCNet). The CCNet collects contextual information in horizontal and vertical directions to enhance pixel-wise representative capability. The CCNet can harvest the contextual information of its surrounding pixels on the criss-cross path through a novel criss-cross attention module for each pixel. By taking a further recurrent operation, each pixel can finally capture the long-range dependencies from all pixels.

It is essential to point out that previous work did not report results with alcohol presence.

This paper aims to segment the iris and pupil from NIR with alcohol included as in shown in Figure \ref{all_stages}. Most of the state-of-the-art algorithms fail in this task.

The article is organized as follows: Section~\ref{relate_work} summarizes the related works on alcohol and semantic segmentation. The database is explained in Section~\ref{databases}. The Eye detection methods proposed are in Section~\ref{eye_det}. The Iris and pupil localization methods are in Section~ \ref{iris-pupil}. The experimental framework and results are presented in Section~\ref{experiments}, and we conclude the article in Section~\ref{conclusions}.

\section{Databases}
\label{databases}

\begin{figure}[H]
\centering
\includegraphics[scale=0.22]{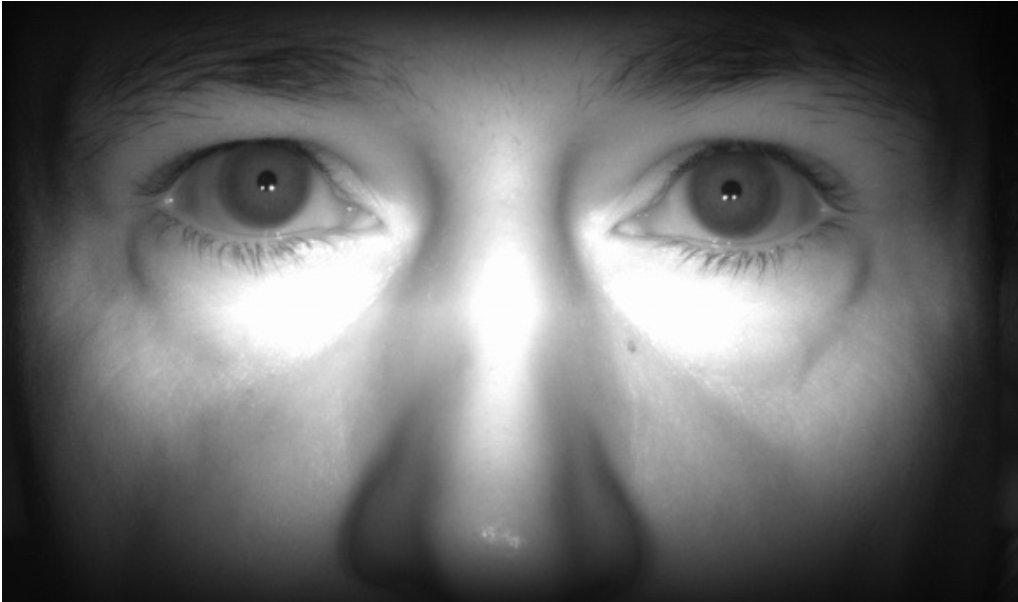}
\includegraphics[scale=0.22]{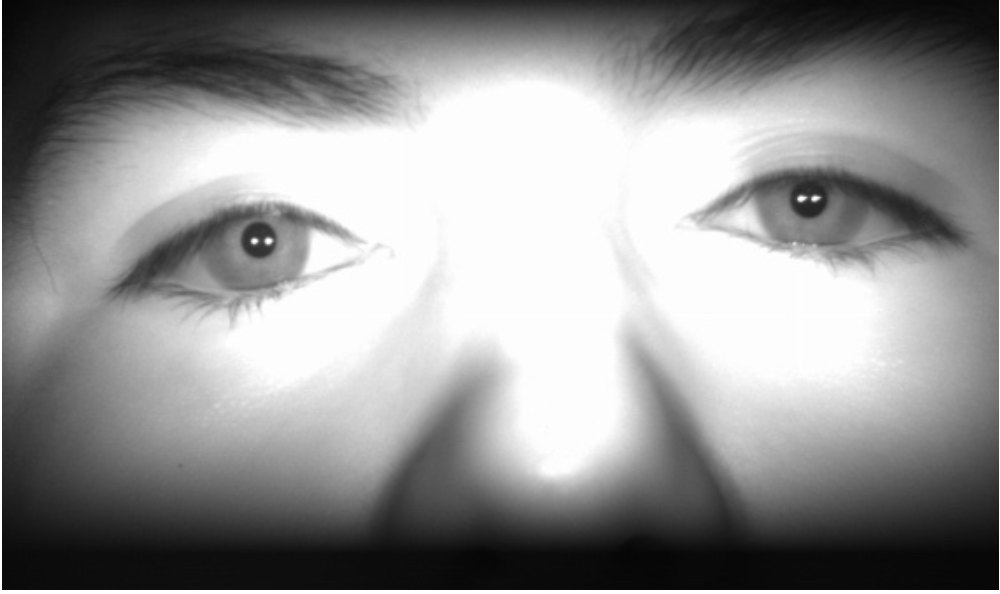}\\
\includegraphics[scale=0.22]{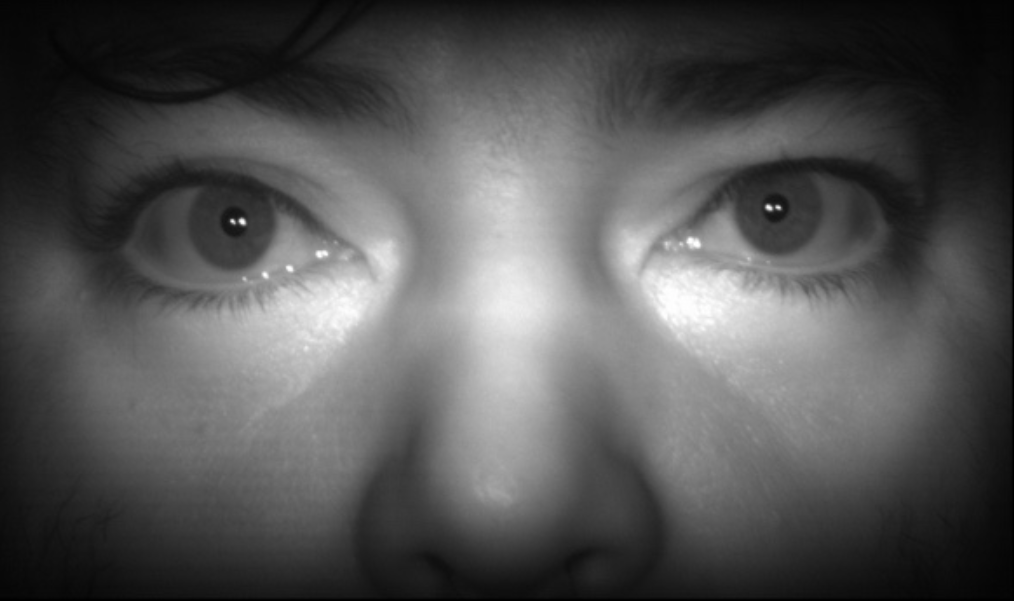}
\includegraphics[scale=0.22]{images/Captura3.png}
\caption{Example of binocular periocular images taken directly from NIR sensor at 30 cm.}
\label{original_capture}
\end{figure}

For this paper, a new database under alcohol presence was created. The images were captured using two different iris NIR sensors: Iritech Gemini, and Iritech Venus \footnote{\url{https://www.iritech.com/products/hardware/gemini-camera}}. Each image has a size of $1280\times760$ pixels. Example of images are shown in Figure \ref{original_capture}.

Two-hundred and sixty-six subjects were included in the experiments under alcohol effect and 765 subjects with no alcohol. Each volunteer was requested to step on a floor mark (30 cm from the camera for Gemini and 50 cm for Venus sensor) and to look at the NIR sensor after having consumed 200 ml of alcohol according to the protocol explained as follows: 

The data capture process was organized in 5 sessions, according to the following protocol:

\begin{itemize}
    \item Session 0: Images were captured when individuals were not under the influence of alcohol.
    \item Session 1: Images were captured 15 minutes after the individual consumed 200 ml of alcohol with a concentration level of 42\%.
    \item Session 2: Images were captured 30 minutes after alcohol consumption (Session 1).
    \item Session 3: Images were captured 45 minutes after alcohol consumption (Session 1).
    \item Session 4: Images were captured 60 minutes after alcohol consumption (Session 1).
\end{itemize}

The room temperature and lighting were controlled and kept constant during the data capturing process. A total of 100 images per eye were captured for each individual per session.

\begin{figure}[H]
\centering
\includegraphics[scale=0.34]{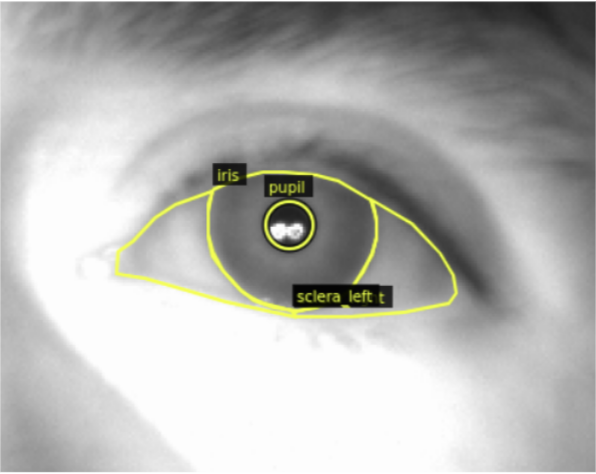}
\includegraphics[scale=0.34]{images/left_label_eye.png}
\caption{\label{manualy_labeled1} Example images with both eyes manually labeled.}
\end{figure}

Overall a total of 21,309 subject-disjoint images were used to train our proposed method. All the images were manually labeled using the VIA tools \cite{viavgg}. For left and right eyes, pupil, iris, and sclera were labeled. This was a very demanding and time-consuming process that took over one year. See Figure \ref{manualy_labeled1}.

After the eye detector was applied to separate left and right eyes, the images were divided into Train (70\%), Validation (20\%), and Test (10\%). That is 14,918 (alcohol: 13,126, and 1,790 No\_alcohol), 4,263 (alcohol: 3.725, and 528 No\_alcohol), 2,134 (alcohol: 1,858, and 256 No\_alcohol) images respectively. The subset separation takes into consideration the amount of real levels of alcohol and non-alcohol cases in a working environment. On average, 10\% of workers present some level of alcohol consumption.

The manually labeled pupils contain pupil radii ranging from 7 to 18 pixels with an average radio of 9 pixels, making both databases complicated for performing segmentation over the pupil.

\begin{figure}[H]
\centering
\begin{tabular}{cc}
\includegraphics [scale=0.235]{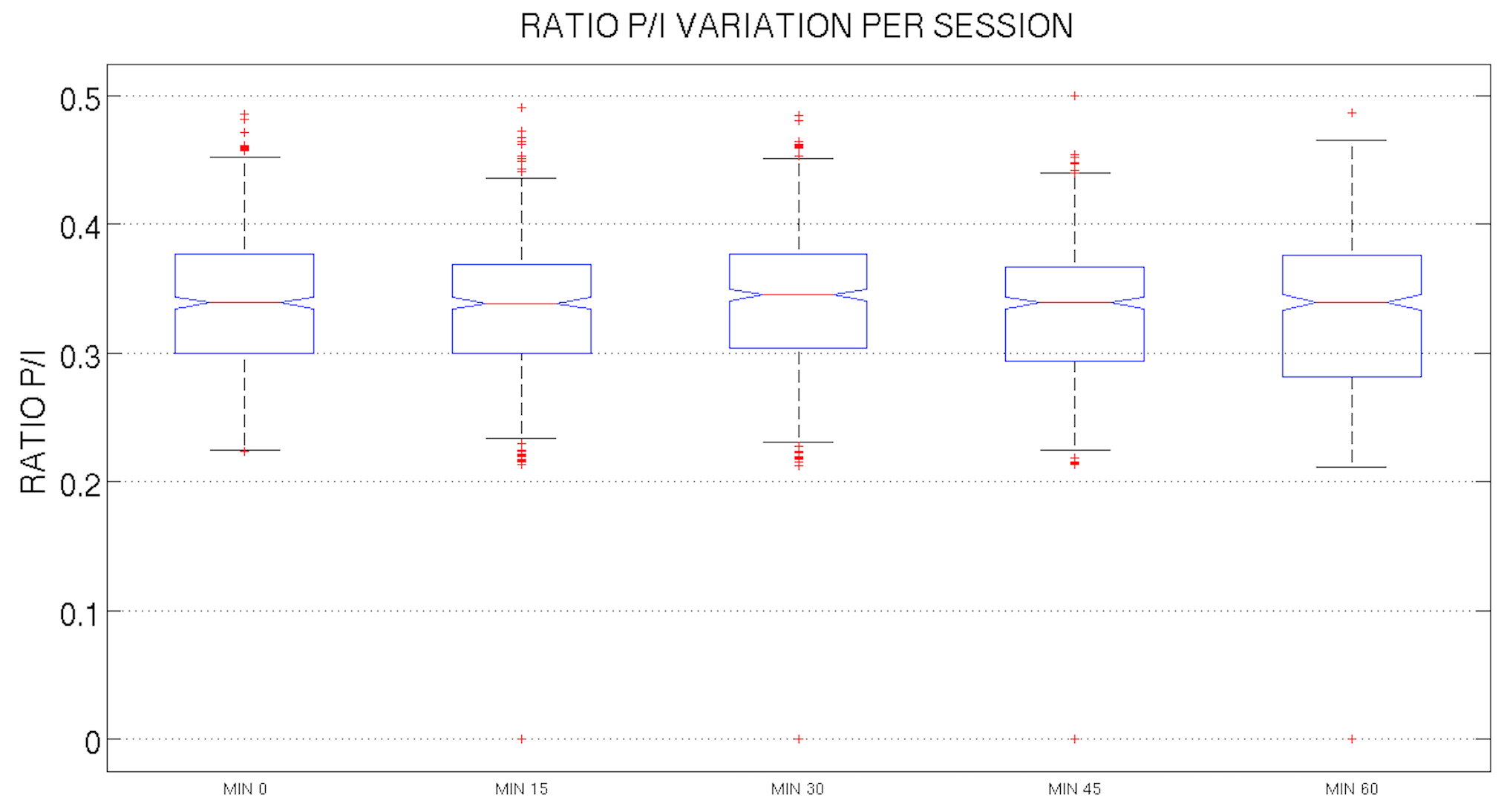} 
\end{tabular}
\caption{\label{box2} Y-axis shows Pupil-Iris Ratio for each session separately. X-axis show the time consumption of alcohol. From left to right: 0, 15, 30, 45 and 60 minutes.}
\end{figure}

Figure \ref{box2} shows a statistical representation of each group present in the dataset according to pupil-iris ratio. In this database, no gender analysis was performed.

As we mentioned before, a stream of images containing both irises were captured - 100 frames in average. This process, which takes five seconds, was repeated for all the participants. During that time, the aperture of the subject’s iris adjusts to NIR light changes of the sensor’s LEDs. This velocity change, measured across all the frames, helps estimate alcohol’s influence on the individuals since alcohol directly affects the velocity of iris adjustment to direct light. 

\subsection{Data augmentation}

\begin{figure}[H]
\centering
\includegraphics[scale=0.28]{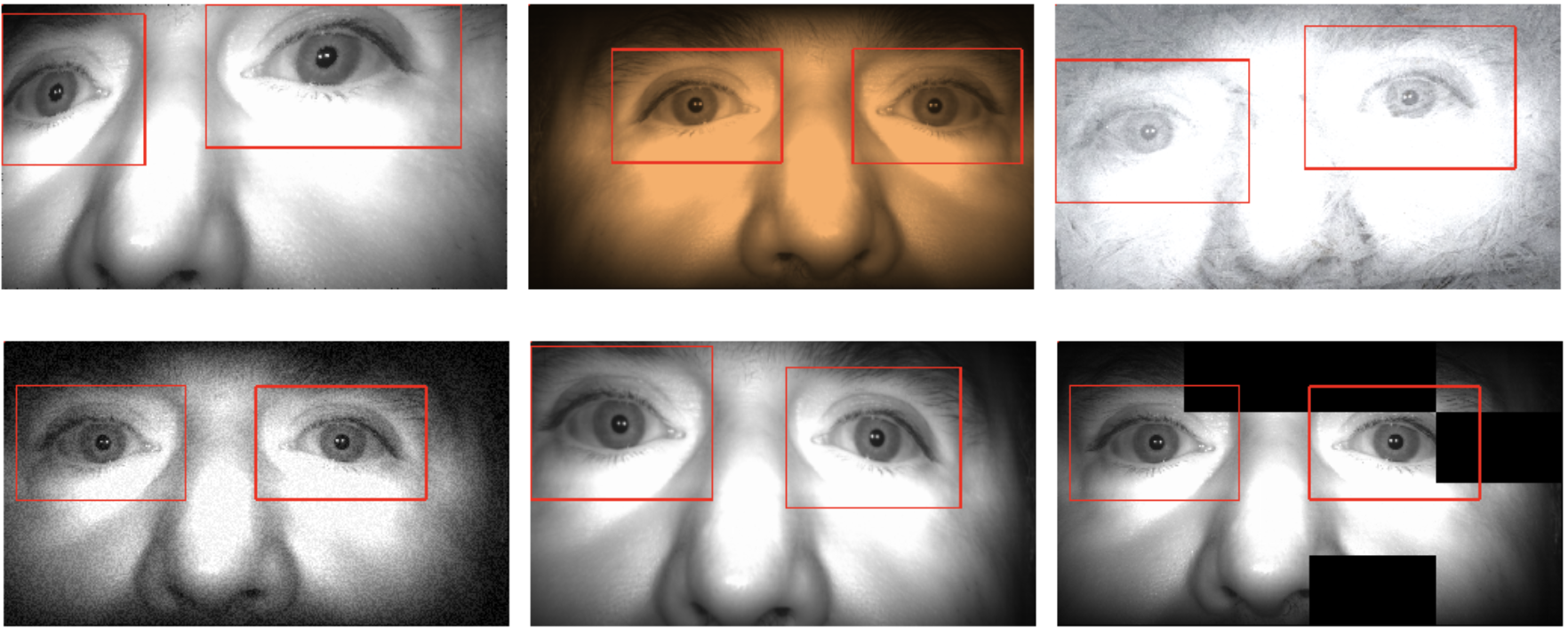}
\caption{Example of aggressive data augmentation from semantic periocular NIR. In red the eye-yolo detection}
\label{DA}
\end{figure}

An aggressive Data Augmentation (DA) was performed using the imgaug library\cite{imgaug}. DA was applied using non-geometrical transformations like adding Gaussian, Laplace, and Poisson noise, as well as image corruption like frost, spatter, snowflakes, and rain effects. Additional non-geometrical effects included adding black patches and adjusting sharpness to the images. Geometrical transformations were also applied. Figure \ref{DA} shows examples of the output images.

\section{Eye Detection in Periocular Images}
\label{eye_det}

\begin{figure}[H]
\centering
\includegraphics[scale=0.40]{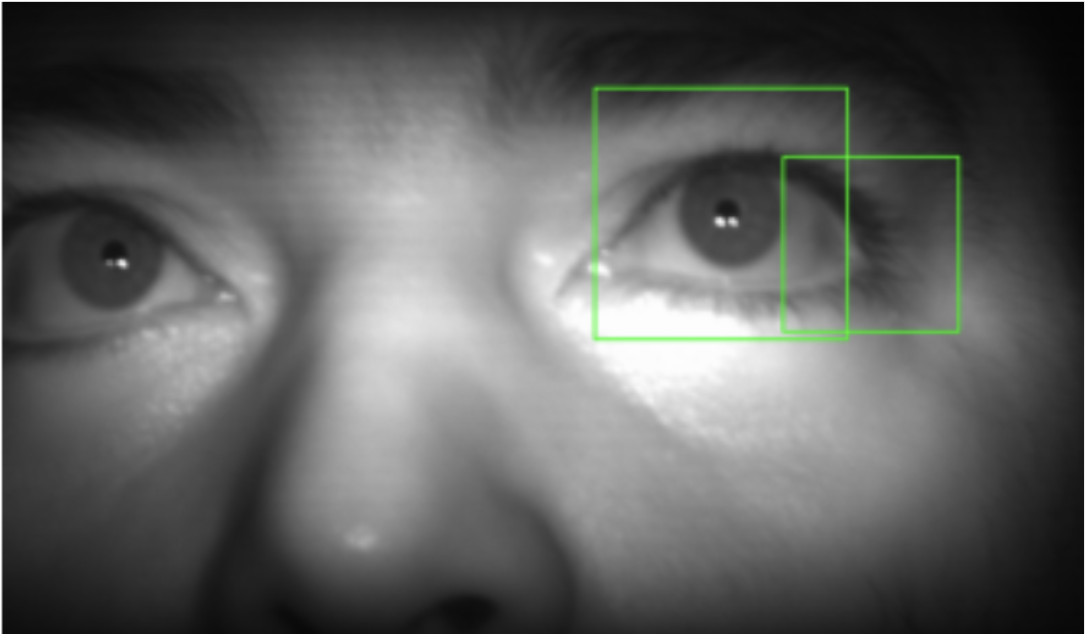}
\caption{\label{haar_fp} Eye detector using Haar-cascade. The green square shows a False positive detection. Also one eye detection is missing.}
\end{figure}

The capture process is cooperative and delivers periocular images with both eyes on it. The images are usually not centralized because of the effect the alcohol had for the volunteer. On average, one hundred frames are captured per subject. This capture process takes five seconds. 
Then, an eye detector was implemented in order to find both eyes in the images and crop them to be segmented in a later step. A traditional eye detector based on haar-cascade \cite{Viola2004haarcascade} was evaluated as a baseline with an 78,60\% of accuracy. However, the results are not reliable and present many false positives. See Figure \ref{haar_fp}. In the end, a new eye detector was trained to reduce the false-positive ratio, called Eye-tiny-yolo.

\subsection{Eye detectors}
You Only Look Once (YOLO) \cite{yolo} is a state-of-the-art detector and predicts multiple bounding boxes per grid cell. At training time, we only want one bounding box predictor to be responsible for each object. We assign one predictor to be “responsible” for predicting an object-based (Eye) prediction with the highest current Intersection Over Union with the ground truth. This metric leads to specialization between the bounding box predictors. Each predictor gets better at predicting specific sizes, aspect ratios, or object classes, improving overall recall. However, the pre-trained tiny-yolo did not work well for our proposed method. This result is because no eye images are used while training it. The proposed Eye-tiny-yolo was retrained from scratch using images of size $416\times416$. The performance improved substantially; however, the two eyes are detected at the same time in bounding boxes of different sizes. Therefore, they are re-sized to $320\times320$ as an input to the segmentation stage.

\begin{figure}[]
\centering
\includegraphics[scale=0.9]{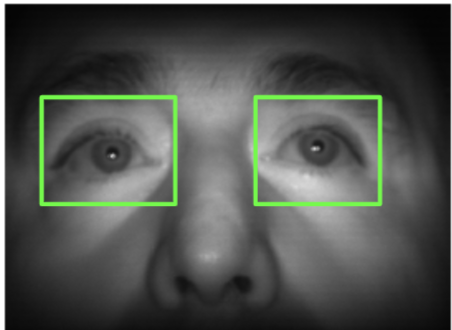}
\caption{\label{res_yolo} Example of Eye-yolo detection with both eyes automatically detected. The nostrils also represent a challenge in this image because the segmenter could have confused it with pupils. The green squares show a correct detection of the left and right eyes.}
\end{figure}

Figure \ref{res_yolo} shows a high confidence detection of both eyes in the same picture at the same time. It is not necessary a double flow and even we can detect the left and the right eyes. This method reached an 98,60\% of accuracy.

\section{Iris and Pupil Localization}
\label{iris-pupil}

\subsection{Mass Center}

\begin{figure*}[]
\centering
\includegraphics[scale=0.62]{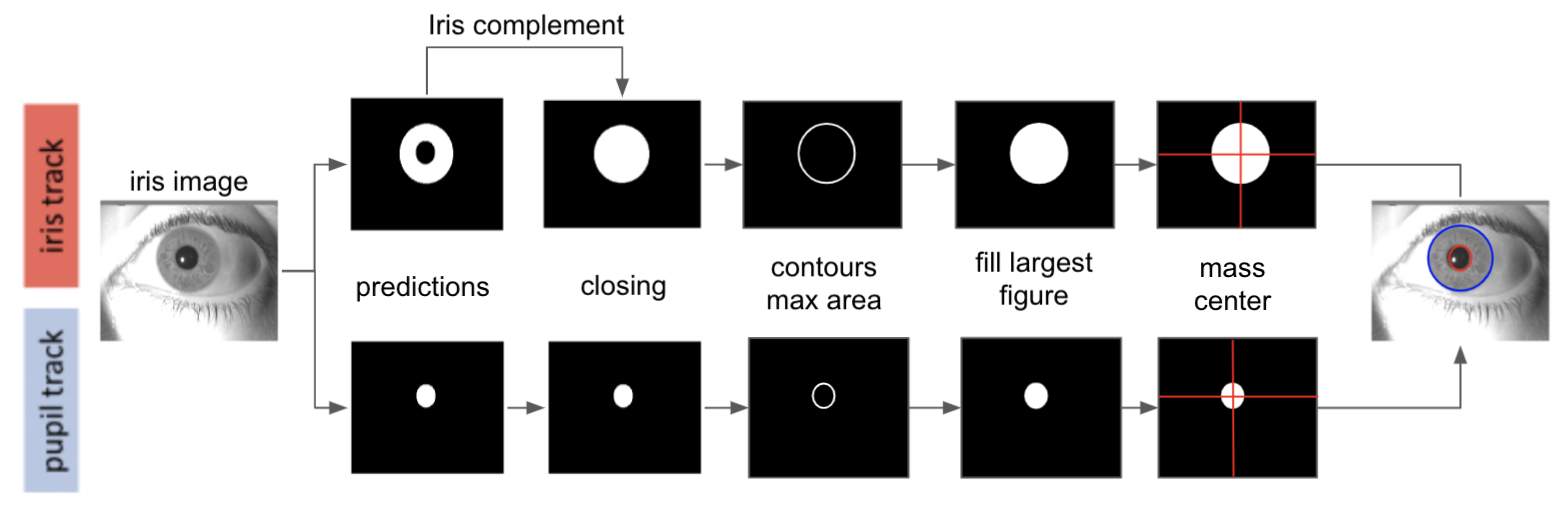}
\caption{Mass Center framework implementation. letters a) Represents the Iris track and b) Represent Pupil track.}
\label{mass_center}
\end{figure*}

This localization method is straightforward and effective. First, a binary image of the pupil and iris is complemented using the $XOR$ operation. As a result, a contour area is obtained for the pupil and iris. Afterward, the most significant area is filled in order to search and estimate the boundaries of the iris and the pupil. This method explores the vertical and horizontal boundaries (edges) of the pixels. With this information, we calculate the radii of the pupil and the iris separately, as shown in Figure \ref{mass_center}.

\subsection{Least-Mean-Squares based Segmentation}
This localization algorithm is conceived to be as lightweight as possible. A binary and morphological operations \cite{Corke17a} were used instead of computationally expensive algorithms that have been applied in traditional approaches, such as Canny edge detection, the Hough transform, and RANSAC \cite{Daugman, Osiris, canny, duda, Fischler}. 
First, A single-channel prediction of the iris region is computed. Then, it was isolated the iris region employing a hole-filling operation \cite{Corke17a}, which removes the pupil. The pupil is then separated by applying a $XOR$ operation between the raw prediction and the iris region. After that, the contour of the iris and the pupil regions was obtained. For this purpose, the $XOR$ operation between the binary image and the erosion of the same binary image was used, thus recovering only the removed pixels during the erosion operation. In the next step, horizontal lines from the iris contour, using Sobel filters \cite{Corke17a} were removed to eliminate eye leads and eyelashes. Finally, the coordinates of the contour pixels and estimation of the best fitting circle using a Least-Mean-Squares (LMS) algorithm \cite{Corke17a} was calculated. As a result, the coordinates of the center and radius of both pupil and iris are obtained. 

\subsection{Mixed Approach}

\begin{figure*}[]
\centering
\includegraphics[scale=0.67]{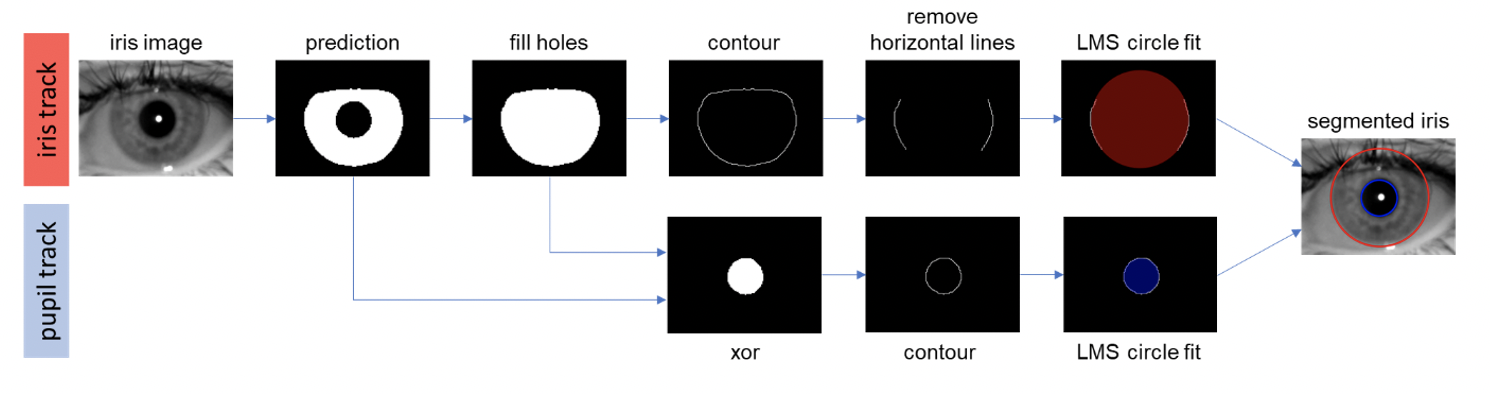}
\caption{\label{tab:centros_lms} Least-Mean-Squares framework implementation.}
\end{figure*}

The "Mixed Algorithm" is a combination of the previous methods. First, the erosion algorithm is applied to find boundaries. After that, the Mass center is used to find the pupil circle. The LMS algorithm is used to find the iris circle. Finally, the Hough transform is applied only when the eyes are partially closed. The combination of these three algorithms allows us to improve the results. See Figure \ref{tab:centros_lms}.

\section{Experiments and Results}
\label{experiments}

\subsection{Metrics}

This section describes all the metrics used to evaluate the whole framework: eye detection, segmentation stage, localization of the iris and pupils. Figure \ref{all_stages} shows the results of the framework with all stages applied.

\subsection{Intersection Over Union}
\begin{figure}[H]
\centering
    \includegraphics[scale=0.25]{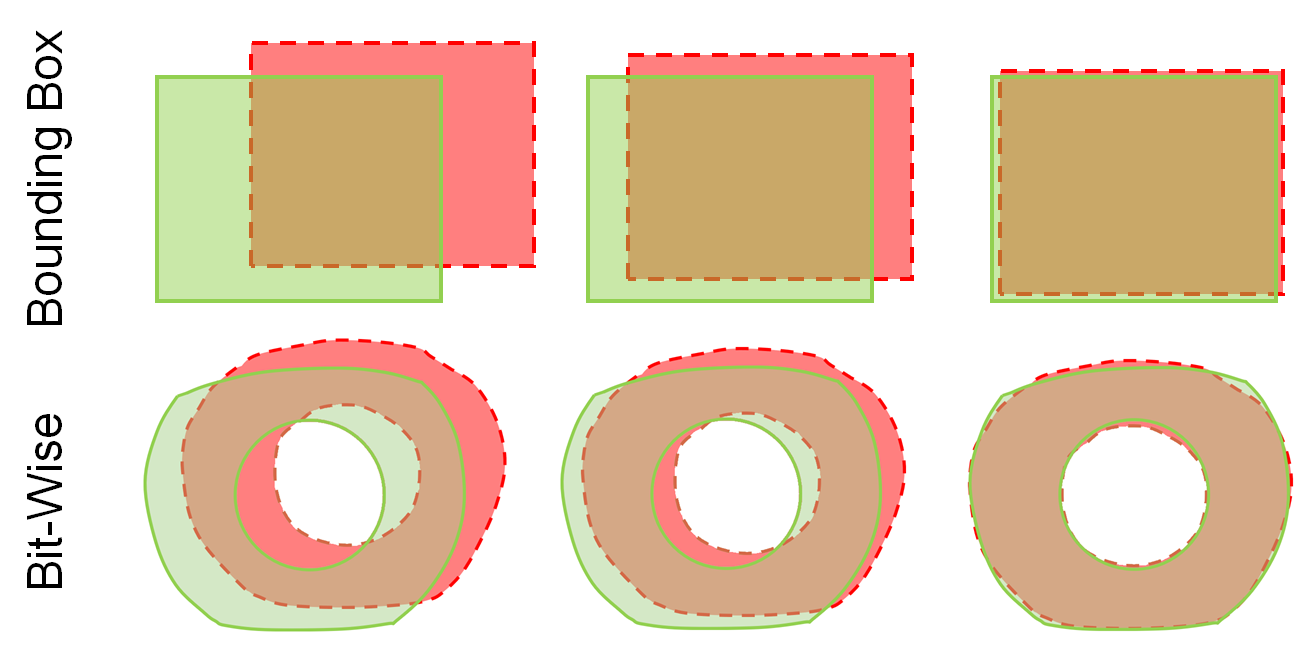}
\caption{\label{fig:IOU2} A visual example of Intersection Over Union and bitwise quality metric. The dashed red box is the predicted detection, the continuous green box is the ground truth, and the gray area is the overlap between the two. The example shows three different IOU scores from left to the right, with the rightmost being the best. Based on \cite{seba}.}
\end{figure}

In order to evaluate eye-tiny-yolo detector, the Intersection over Union (IoU) metric was used. The IoU measures the overlap between two boundary images. This is used to measure how much the boundary predicted by the algorithm overlaps with the ground truth (the real object). Traditionally, state-of-the-art datasets reported an IoU threshold equal to or greater than 0.5. Figure \ref{fig:IOU2} show the graphical effect to apply IoU. The IoU is calculated using the following equation: 

\begin{equation}\label{eq_iou}
    IoU = \frac{A \cap B}{ A \cup B}
\end{equation}

On the other hand, for the evaluation of the segmentation networks we use the bit-wise version of IoU. This IoU assesses how much of the predicted binary mask correlates with the ground-truth mask pixel by pixel. The following equation is used for its calculation:

\begin{equation}\label{eq_iou2}
    IoU = \frac{\sum and(A, B)}{\sum or(A, B)+c} ,
\end{equation}

where A and B are binary images, and c is a small constant that prevents the IoU from taking an infinite value when there is no overlap between A and B.

\subsection{Experiment 1:Osiris}
\begin{figure}[H]
\centering
\includegraphics[scale=0.16]{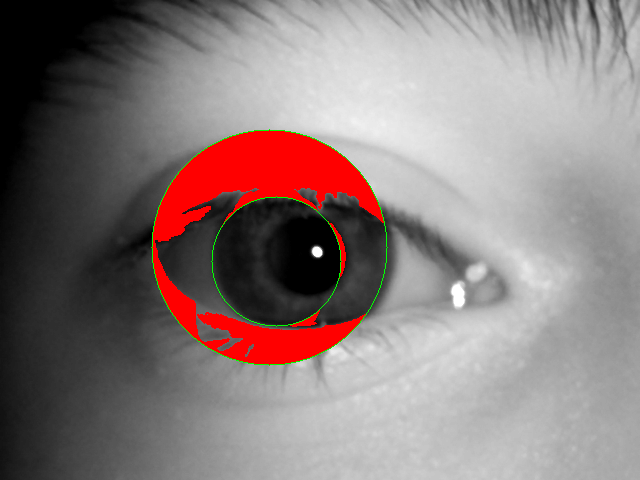}
\includegraphics[scale=0.16]{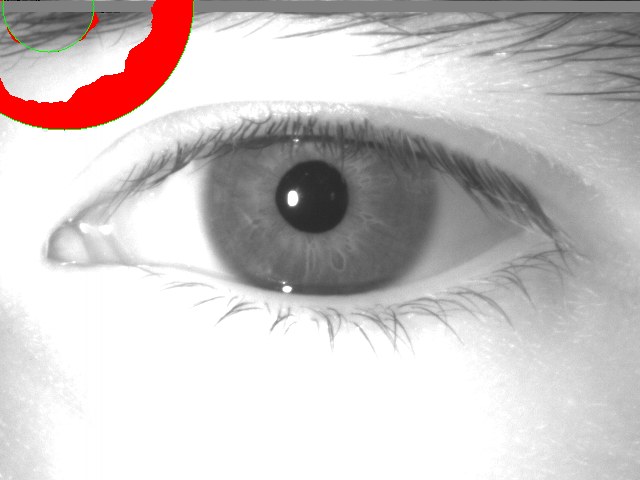}\\
\includegraphics[scale=0.16]{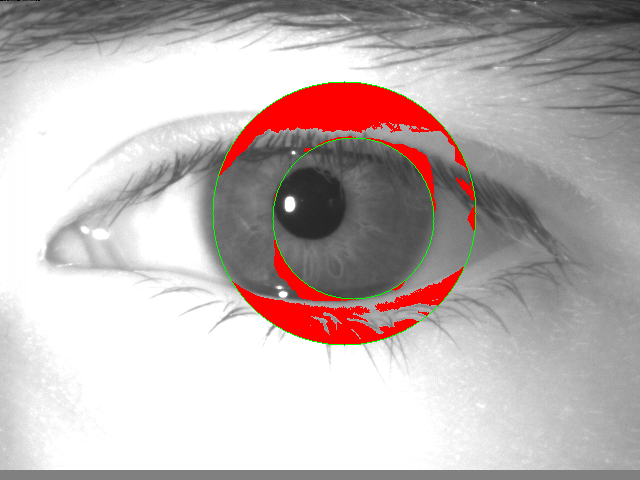}
\includegraphics[scale=0.16]{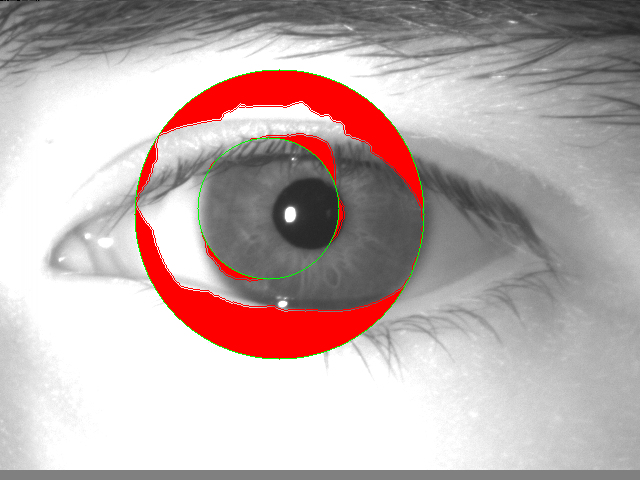}

\caption{Example of wrong segmentation under presence of alcohol for traditional Osiris software.}
\label{osiris_error}
\end{figure}

As a baseline, the Osiris software was used to segment the pupil and the iris from alcohol images. However, tthe resulting segmentation performance was low. The alcohol images show the size of pupil very large or small compared to traditional images in normal condition. Therefore, the default parameters are not valuable for the proper segmentation of the image. Figure \ref{osiris_error} Show examples of wrong segmentation for alcohol images. The method fails in both pupil and iris localization.

\subsection{Experiment 2:DeepVOG}

Two experiments were developed for DeepVOG. For the first one, pre-trained models were used. For the second one, a DeepVOG was trained from scratch.  
The DeepVOG was trained using an Adam optimizer with a learning rate of $1e-5$, a Tversky Loss, trained during 400 epochs, a batch size equals 8 with a step size of 64 and an input shape of $320\times 320\times3$.
The modified DeepVOG was also tested with the same database to check their performance under a complex scenario (very low pupil size radio). As we mentioned before, the database was manually labeled. The coordinate allows us to crop the eyes from the periocular images manually or automatically using Eye-tiny-yolo.

For the manually cropped images, they were cropped, splitting the $1280\times760$ image into two images of $640\times480$ pixels; each one of them has one eye. The second cropping was made using Eye-tiny-yolo. The output image was resized to its original size and placed again in a black image of $1280\times760$ to compute the IoU. Table \ref{tab:centro1} shows the results for DeepVOG and DeepVOG2.

\subsection{Experiment 3:DenseNet10}

One of the goals of this work is to reduce the layers and the concatenation matrix's size from the original implementation of DenseNet56 and DenseNet101. For doing so, a feature extractor and two paths (Down-sampling and one Up-sampling) were modified. The down-sampling path has 1 Transition Down (TD), and an up\-sampling course has 1 Transition Up (TU) instead of the four Transitions $(2TD+2TU)$ used in the traditional approach. For each layer $i$, the number of feature maps $k$ obtained is given by the following equation: $k0+k\times (i−1$), where $k0$ is the number of channels in the input layer.
In order to define the best number of filters $k$ to use in each layer, a grid search from $k=3$ to $k=15$ was used. The best result was obtained with $k=3$. The following parameters were used to train our DenseNet10: Epochs: $200$, Batch Size: $6$, Optimizer: RMSprop,LR: $1e-4$, Decay: $1e-6$, Dropout: $0.15$, with aggressive Data Augmentation and an Input shape: $320\times320\times3$. The resulting architecture can be trained with a Batch Size of $32$. All the experiments were ran using a GPU\-1080ti with 32GB RAM.

\begin{figure}[]
\centering
\includegraphics[scale=0.33]{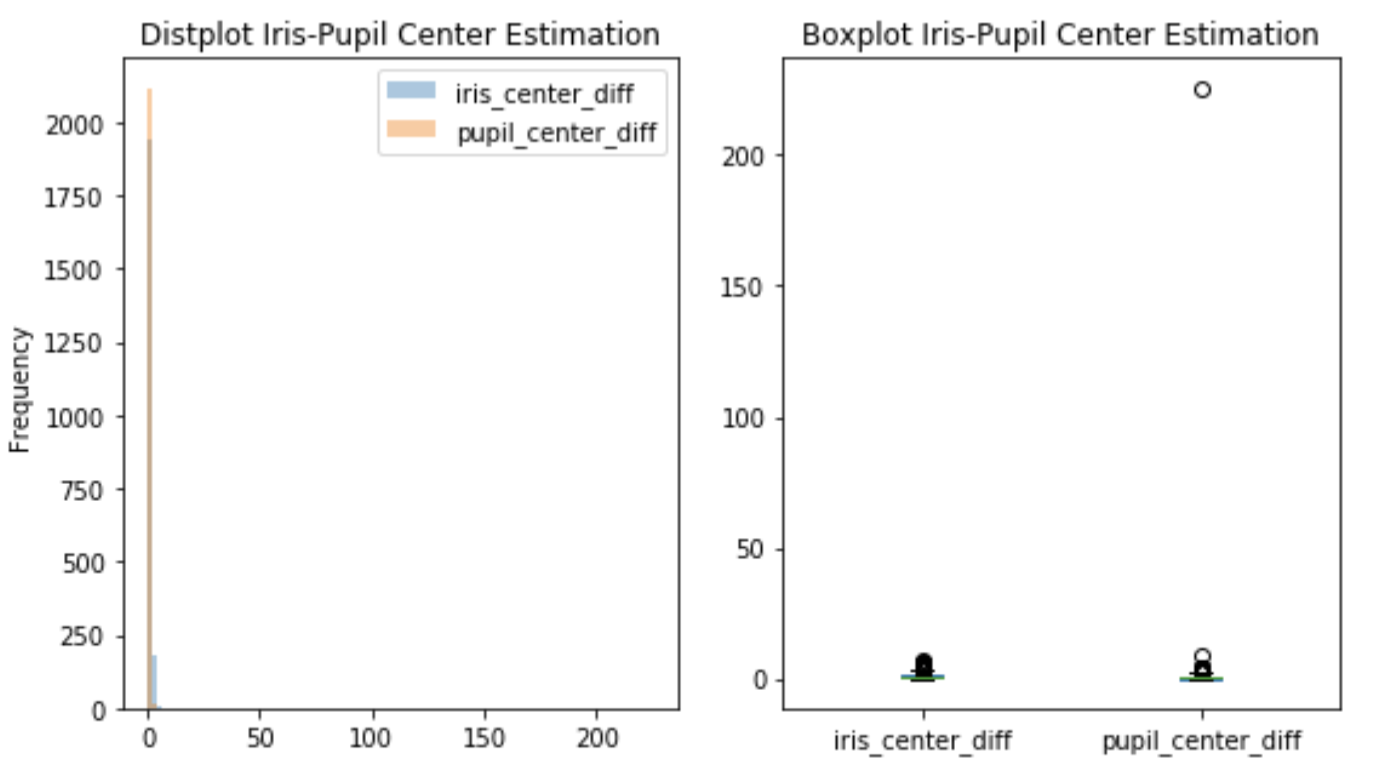}
\caption{\label{densenet1} Histogram of the distribution error for the pupil and the iris center localization using DenseNet10.}
\end{figure}

Figure \ref{densenet1} shows a histogram with the distribution error for pupil and iris localization with images under alcohol presence using the DenseNet10 method. The left figure shows a histogram error rate of 1 pixel for the test set. The right figure shows the plot distribution for the iris and pupil.

\subsection{Experiment 4:CCNet}

Criss-Cross attention network (CCNet) was used in this approach to get an efficient model with a low number of parameters. Our model was trained from scratch with the presented dataset. The input image used is $360\times240\times1$. The best parameters used were: Batch size: $60$, LR: $1e-3$, Epochs: $200$. This network was trained and tested with a 6GB GTX-1660 GPU and 16 GB of RAM. 

\begin{figure}[H]
\centering
\includegraphics[scale=0.33]{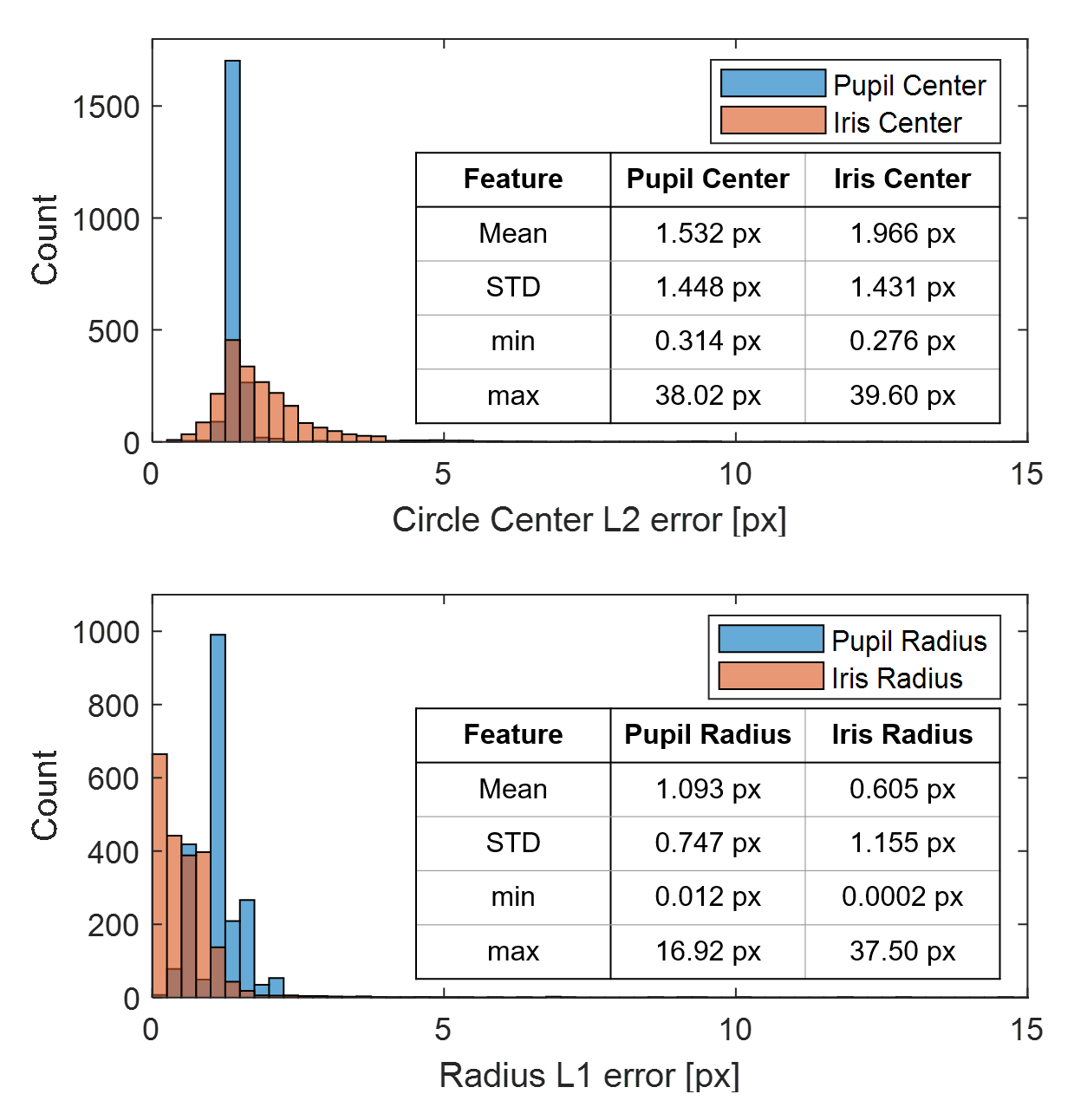}
\caption{\label{ccnet} Histogram of the distribution error for pupil and iris center localization using CCNET.}
\end{figure}

Figure \ref{ccnet} shows a histogram with the distribution error for pupil and iris localization with images under alcohol presence using CCNet model and the LMS algorithm. 

Table \ref{tab:centro1} shows the comparison results among the Osiris as a baseline, DeepVOG, Modified DeepVOG2, CCNET, and DenseNet10. The Osiris system performs over 63.4 \% Mean IoU, while the Original DeepVOG obtained a 90.03\%. DeepVog2 trained from scratch to reach 91.30\%. The proposed method trained under the database alcohol presents a 94,58 \% of Mean IoU.

\begin{table}[H]
\centering
\caption{\label{tab:centro1} Summary of the Segmentation methods under alcohol influences.}
\begin{tabular}{|c|c|c|c|}
\hline
\textbf{Method} & \textbf{Mean IoU} & \textbf{Std} & \textbf{Nº Param} \\ \hline
Osiris          & 0.6336       & 0.2780        & N/A                \\ \hline
DeepVog         & 0.9003       & 0.0710        & 2,058,979                \\ \hline
DeepVog2        & 0.9130        & 0.0170        & 2,058,979                \\ \hline
DenseNet10      & 0.9458        & 0.0160         & 210,732                \\ \hline
CCNET           & 0.9190        & 0.0400         & 122.514                 \\ \hline
\end{tabular}
\end{table}

Table \ref{tab:centros2} shows the results for the five different metrics. Traditional Hough and Mass center and three proposed metrics to measure the center of the pupil and radii.

\begin{table*}[]
\centering
\caption{\label{tab:centros2} Comparison metrics for all the methods to estimate the localization of the iris and pupil using CCNet. The table shows the error in pixels and the standard deviation. The best results are highlighted in bold.}

\begin{tabular}{|c|c|c|c|c|}
\hline
\textbf{Metric}                     & \textbf{Hough}        & \textbf{Mass Center}      & \textbf{LMS}          & \textbf{Mixed} \\ \hline
Pupil Center (L2 error)             & 3.9 $+/-$ 4.58        & \textbf{1.43 $+/-$ 0.11}  & 1.86 $+/-$ 3.77       & 1.53 $+/-$ 1.45    \\ \hline
Iris Center (L2 error)              & 4.6 $+/-$ 4.42        & 7.30 $+/-$ 5.43           & 2.28 $+/-$ 300        & \textbf{1.96 $+/-$ 1.43}    \\ \hline
Pupil Radius on x (L1 error)        & 2.28 $+/-$ 4.28       & 1.96 $+/-$ 0.36           & 1.17 $+/-$ 1.02       & \textbf{1.09 $+/-$ 0.75}    \\ \hline
Pupil Radius on y (L1 error)        & 2.34 $+/-$ 3.80       & 1.64 $+/-$ 0.57           & 1.19 $+/-$ 2.24       & \textbf{1.12 $+/-$ 1.99 }   \\ \hline
Iris Radius on x (L1 error)         & 2.67 $+/-$ 4.09       & 3.03 $+/-$ 1.12           & 0.72 $+/-$ 1.50       & \textbf{0.604 $+/-$ 1.16}   \\ \hline
Iris Radius on y (L1 error)         & 14.53 $+/-$ 7.16      & 4.27 $+/-$ 1.35           & 0.16 $+/-$ 0.29       & \textbf{0.157 $+/-$ 0.289}  \\ \hline
\end{tabular}
\end{table*}

\begin{figure*}[]
\centering
\includegraphics[scale=0.52]{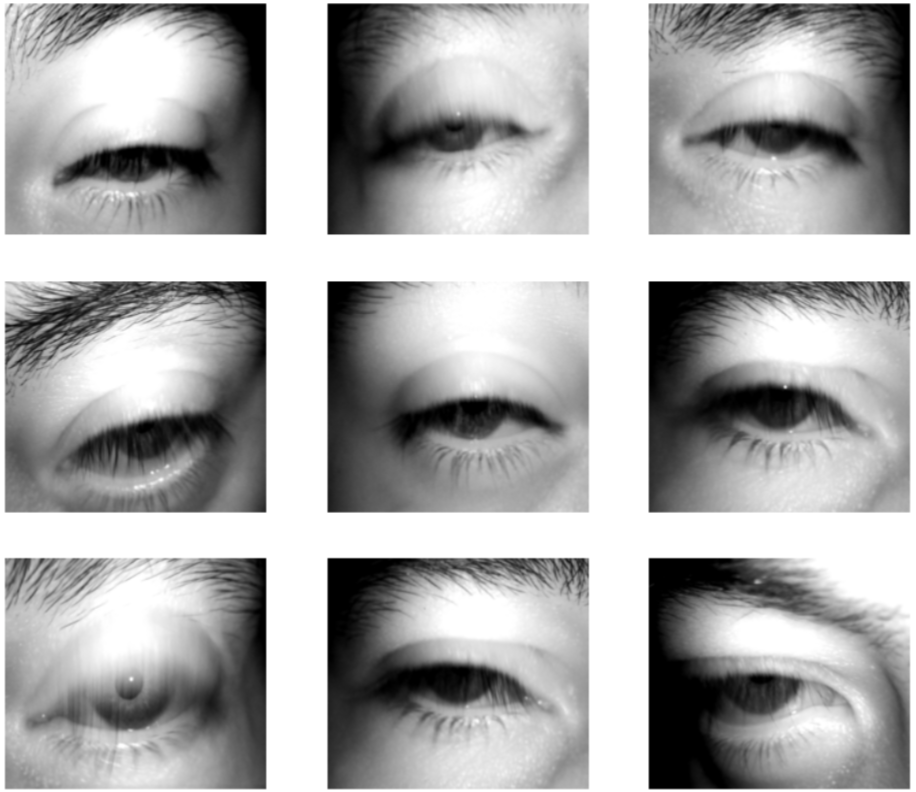}
\includegraphics[scale=0.52]{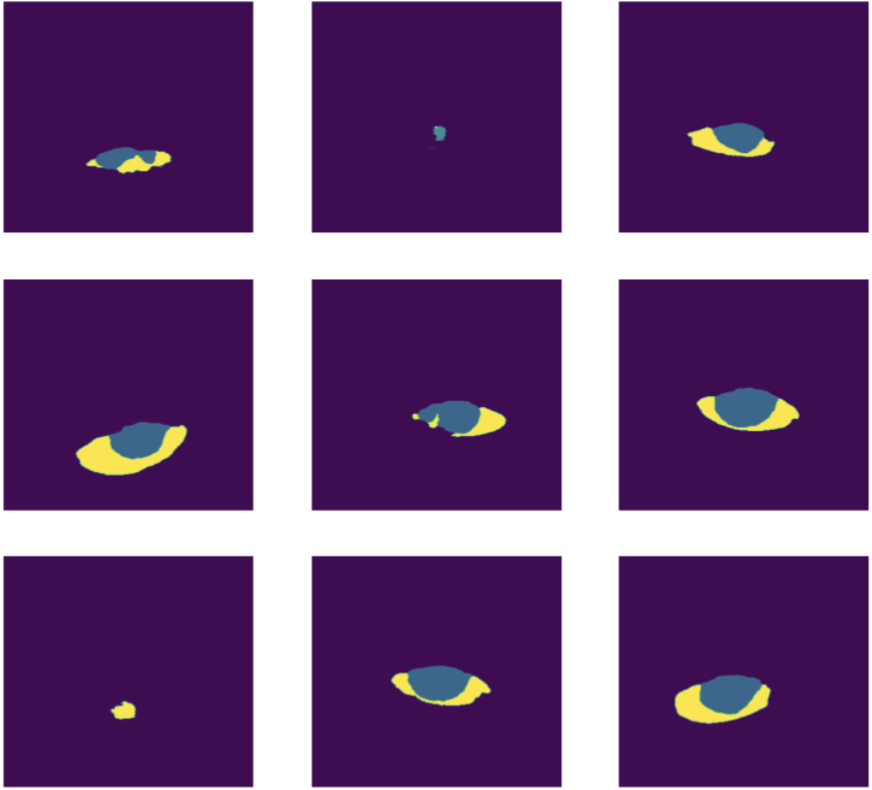}
\caption{\label{sumamry_segemtation1} Example of semi-closed challenging images. Top original images under alcohol consumption. Bottom: DeepVog2 results.}
\end{figure*}

\begin{figure*}[]
\centering
\includegraphics[scale=0.52]{images/original_image.png}
\includegraphics[scale=0.52]{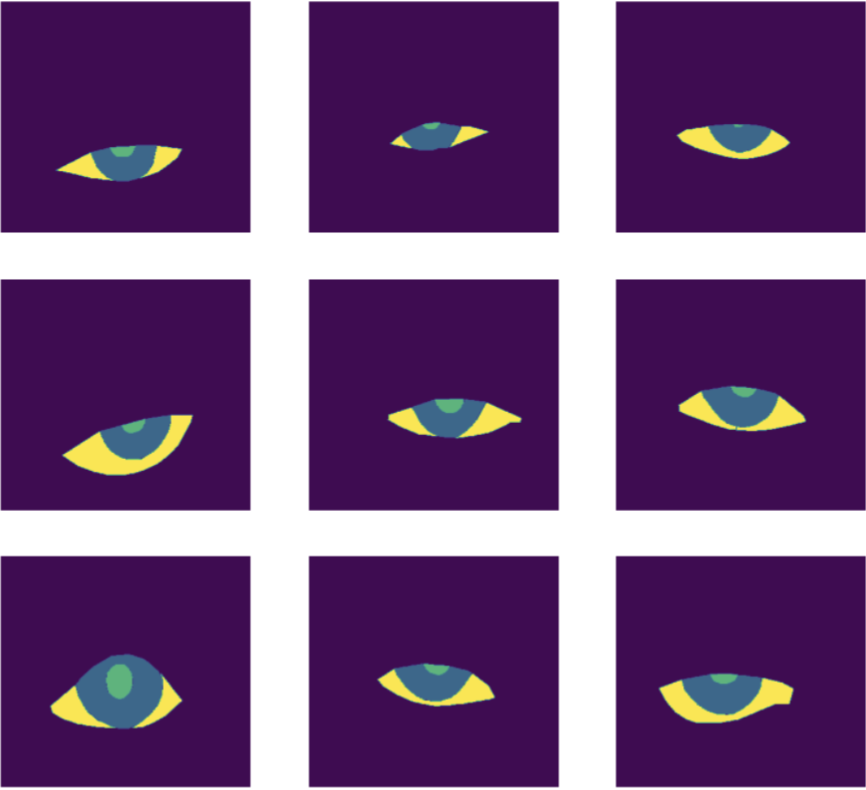}
\caption{\label{sumamry_segemtation2} Example of semi-closed challenging images. Top original images under alcohol consumption. Bottom: Our proposed DenseNet10 results.}
\end{figure*}

Figure \ref{sumamry_segemtation1} and Figure \ref{sumamry_segemtation2} show a set of challenging images with semi-closed eyes under alcohol effects. Left: The images in grayscale show semi-closed eyes where the pupil is not visible and presents some eyelashes occlusions. Figure \ref{sumamry_segemtation1} Right: figures show the results for DeepVOG2. This method reaches good results, but it cannot deal very well with semi-closed eyes. Figure \ref{sumamry_segemtation2} right, the DenseNet10 method shows it can detect these semi-closed eyes with high precision.

\subsection{Experiment 5: Grand-Mean}
The grand mean \cite{RoehrsandRoth} of a set of multiple subsamples is the mean of all observations: each data point, divided by the pooled sample size. This analysis type is used to analyze the differences in the data series acquired between different trials. The pooled mean allows to analyze the general behavior of the studied group and therefore provides guidelines of the "waveform" for each group. Grand mean is widely used in the processing of EEG signals, especially in cognitive evoked potentials, since it allows to obtain the fundamental component of the EEG signal using individual signals that by themselves can be very noisy \cite{eeg2}. In this way, the grand mean enables us to estimate the tendency. This tendency capacity was used to analyze the measures obtained from alcohol and no-alcohol images for each subject using the pupil and iris radii.

\begin{figure}[H]
\centering
\begin{tabular}{cc}
\includegraphics [scale=0.45]{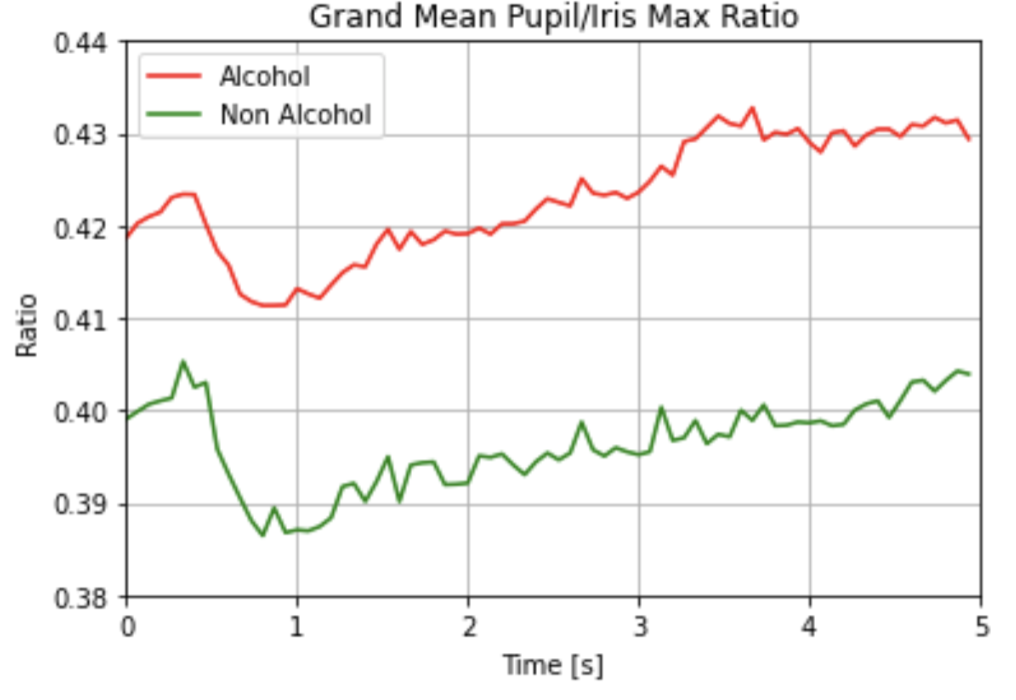} 
\end{tabular}
\caption{\label{grandmean} Grand mean Curves of the alcohol (Red) and no alcohol (Green) population.}
\end{figure}

Figure \ref{grandmean} shows the population behavioral study as such a time sequence of 5 seconds of pupil size recordings. This analysis was conducted to determine whether there were differences in pupil radius size under alcohol and non-alcohol conditions.

In order to obtain the alcohol and non-alcohol curves, the grand mean was estimated for the pupil radii at each time instant for all the subjects in each one of the groups. Thus, it is possible to define the baseline behavior curve for people in the presence and absence of alcohol consumption. The analysis shows a difference in pupil size's temporal behavior, with larger size (more significant dilation) in subjects under the effects of alcohol. It is important to note that this analysis shows the average behavior of the population, so it is not possible to use it as a single variable to separate both groups.

\section{Conclusion}
\label{conclusions}

According to the results, it is possible to measure the alcohol presence and changes in the behavior using the frame of the eyes. The diameters of the pupil and iris present abnormal sizes that confuse traditional approaches and do not allow us to use a parametric method such as Osiris. The semi-closed eyes and eyelashes present a real challenge to efficiency and high accuracy about IoU. A mixed localization method is more suitable for the measure; nevertheless, we reach a high precision of less than one pixel. The number of parameters for the best semantic segmentation approach is also relevant because it allows us to implement this framework in a mobile device or commercial hardware. The proposed system involves two efficient methods: CCNet and DenseNet. DenseNet10 obtained a high score but a higher number of parameters in comparison to CCNet. The database capture for this project is also a significant contribution. This database will be available for a research community by the end of 2022. As future work, we need to study the feasibility of measuring fitness for duties using a regular NIR iris sensor. The results obtained show that it is possible to accomplish this goal.

\section*{Acknowledgment}
This work is fully supported by the Agencia Nacional de Investigacion y Desarrollo (ANID) throught FONDEF IDEA Nº ID19I10118 leading by Juan Tapia Farias - DIMEC-UChile. And the collaboration of the German Federal Ministry of Education and Research and the Hessen State Ministry for Higher Education, Research and the Arts within their joint support of the National Research Center for Applied Cybersecurity ATHENE.

Further thanks to all the volunteers that participated in this research.

\ifCLASSOPTIONcaptionsoff
  \newpage
\fi

\bibliographystyle{IEEEtran}
\bibliography{mybibfile.bib}

\begin{IEEEbiography}[{\includegraphics[width=1in,height=1.25in,clip,keepaspectratio]{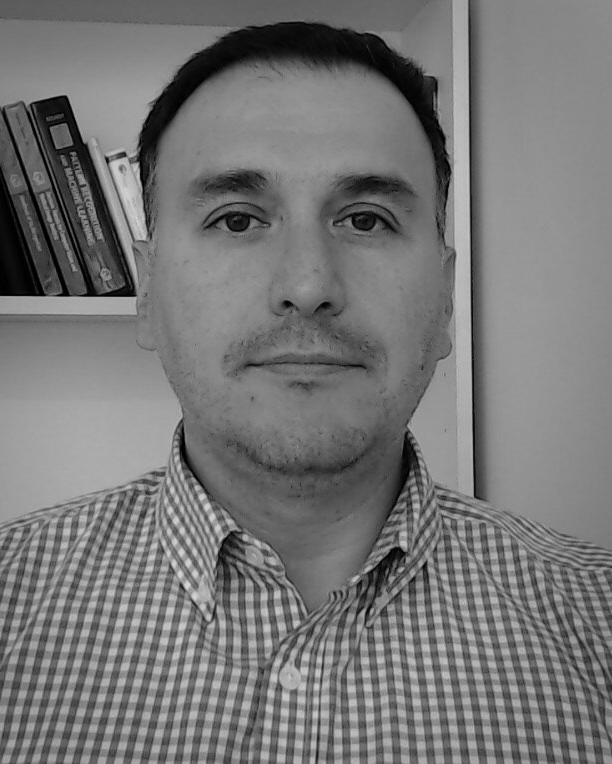}}]{Juan Tapia} received a P.E. degree in Electronics Engineering from Universidad Mayor in 2004, a M.S. in Electrical Engineering from Universidad de Chile in 2012, and a Ph.D. from the Department of Electrical Engineering, Universidad de Chile in 2016. In addition, he spent one year of internship at University of Notre Dame (USA). In 2016, he received the award for best Ph.D. thesis. From 2016 to 2017, he was an Assistant Professor at Universidad Andres Bello. From 2018 to 2020, he was the R\&D Director for the area of Electricity and Electronics at Universidad Tecnologica de Chile - INACAP. He is currently a Senior Researcher at Hoschule Darmstadt(HDA), and R\&D Director of TOC Biometrics. His main research interests include pattern recognition and deep learning applied to iris/face biometrics, vulnerability analysis, morphing, feature fusion, and feature selection.
\end{IEEEbiography}

\begin{IEEEbiography}[{\includegraphics[width=1.10in,height=1.25in,clip,keepaspectratio]{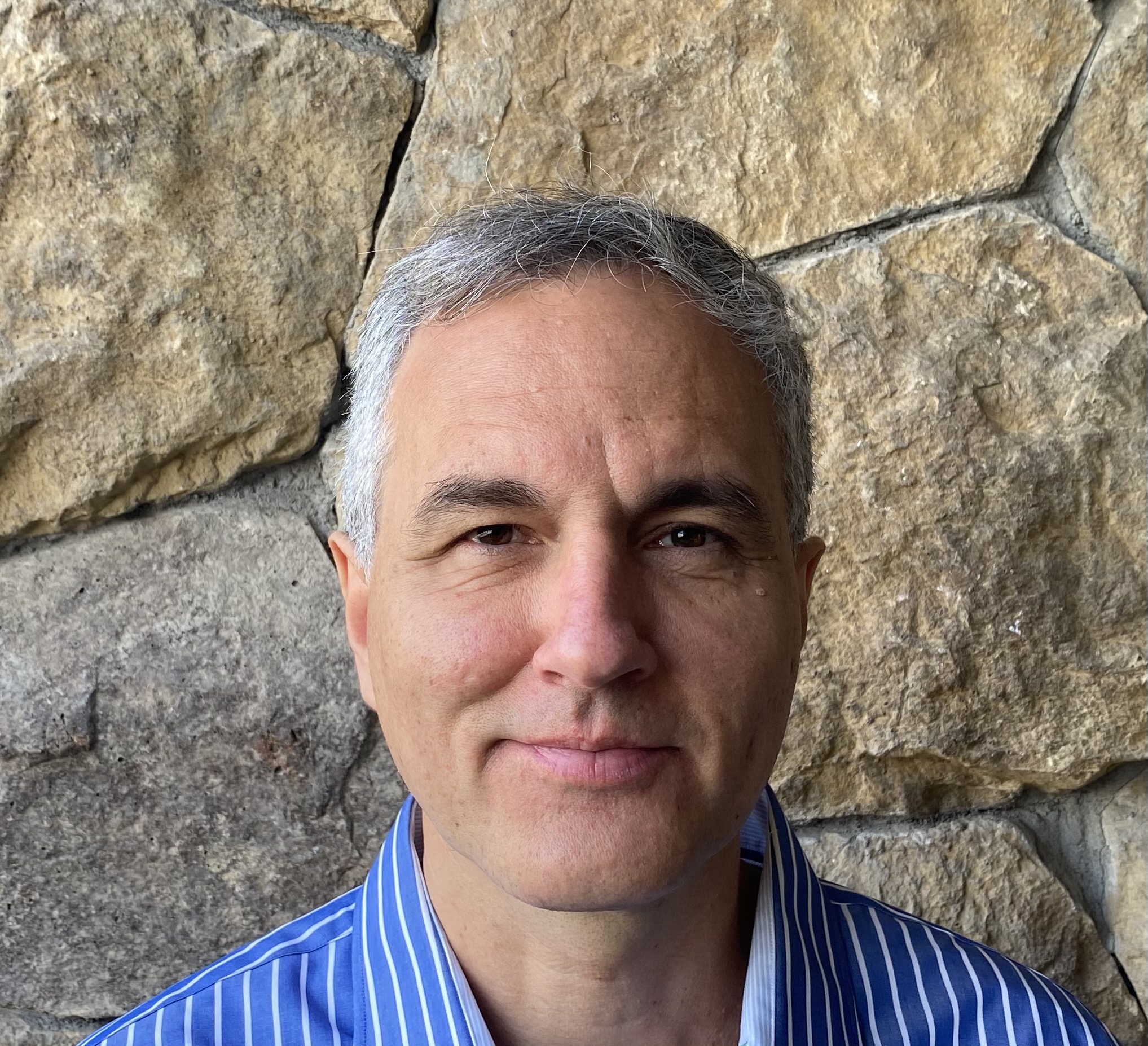}}]{Enrique Lopez Droguett} Droguett is Professor in the Civil \& Environmental Engineering Department and the Garrick Institute for the Risk Sciences at the University of California, Los Angeles (UCLA), USA, and Associate Editor for both the Journal of Risk and Reliability, and the International Journal of Reliability and Safety. He also serves in the Board of Directors of the International Association for Probabilistic Safety Assessment and Management (IAPSAM). Prof. Lopez Droguett conducts research on Bayesian inference and artificial intelligence supported digital twins and prognostics and health management based on physics informed deep learning for reliability, risk, and safety assessment of structural and mechanical systems. His most recent focus has been on quantum computing and quantum machine learning for developing solutions for risk and reliability quantification and energy efficiency of complex systems, particularly those involved in renewable energy production. He has led many major studies on these topics for a broad range of industries, including oil and gas, nuclear energy, defense, civil aviation, mining, renewable and hydro energy production and distribution networks. Lopez Droguett has authored more than 250 papers in archival journals and conference proceedings.
\end{IEEEbiography}

\begin{IEEEbiography}[{\includegraphics[width=1in,height=1.25in,clip,keepaspectratio]{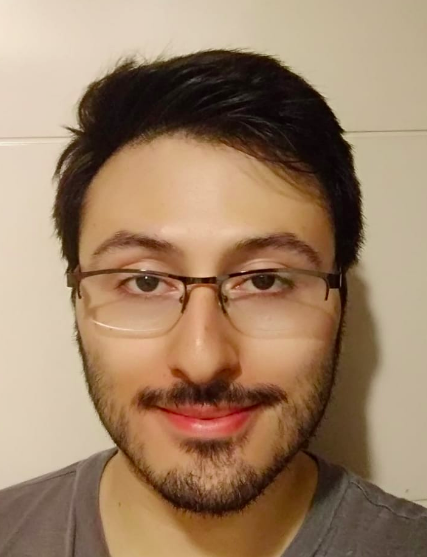}}]{Andres Valenzuela} received a B.S. in Computer Engineering from Universidad Andres Bello, Faculty of Engineering in Santiago, Chile in 2020. His main interests include computer vision, pattern recognition and Deep learning applied to semantic segmentation problems, focusing in NIR and RGB eyes images. Currently, he is working as a researcher at Universidad de Chile and TOC biometrics-Chile.
\end{IEEEbiography}

\begin{IEEEbiography}[{\includegraphics[width=1in,height=1.35in,clip,keepaspectratio]{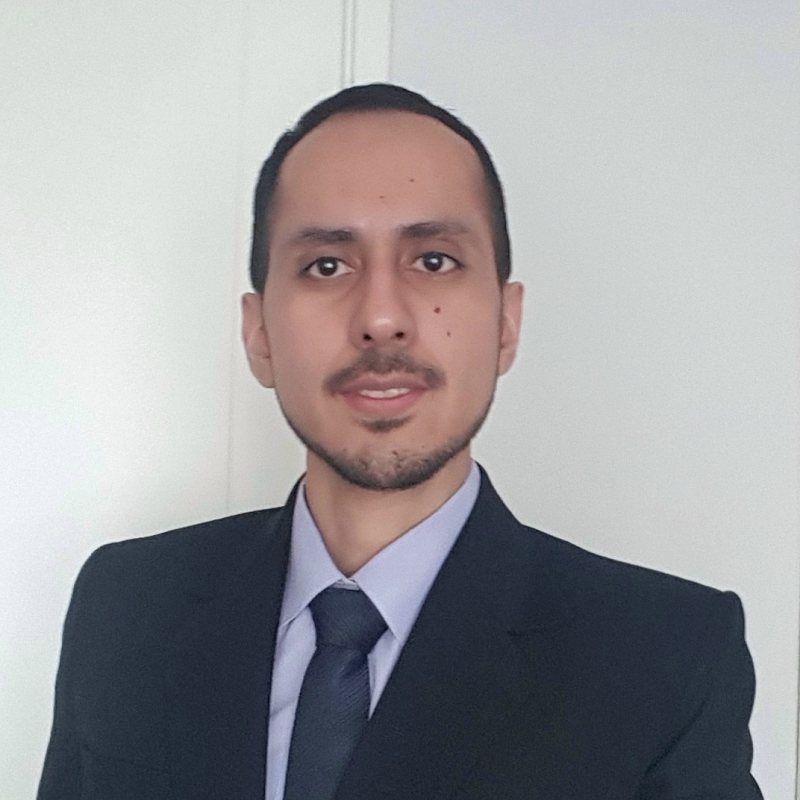}}]
{DANIEL P. BENALCAZAR} (M ‘09) was born in Quito, Ecuador in 1987. He obtained the B.S. in Electronics and Control Engineering from Escuela Politecnica Nacional, Quito, Ecuador in 2012. He received the M.S. in Electrical Engineering from The University of Queensland, Brisbane, Australia in 2014 with a minor in Biomedical Engineering. He obtained the Ph.D. in Electrical Engineering from Universidad de Chile, Santiago, Chile in 2020. From 2015 to 2016, he worked as a Professor at the Central University of Ecuador. Ever since, he has participated in various research projects in biomedical engineering and biometrics. Mr. Benalcazar working as a researcher at Universidad de Chile and TOC biometrics-Chile.
\end{IEEEbiography}

\begin{IEEEbiography}[{\includegraphics[width=1in,height=1.25in,clip,keepaspectratio]{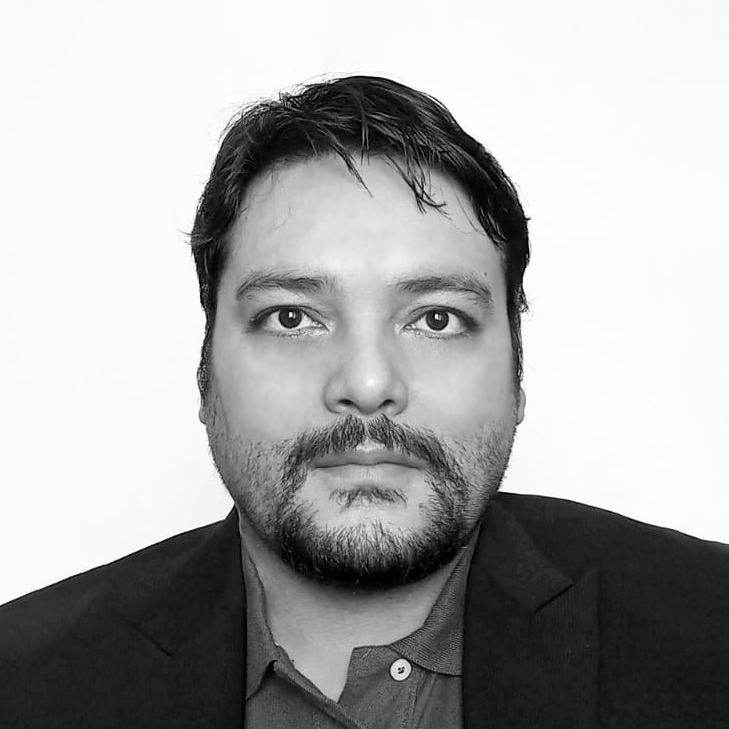}}]{Leonardo Causa} received the P.E. degree in electrical engineering from the Universidad de Chile in 2012, and the M.S. degree in biomedical engineering (BME) from the Universidad de Chile in 2012, he is also finishing the Ph.D. degree in Electrical Engineering and Medical Informatics by cotutelage from U. de Chile and Universite Claude Bernard Lyon 1. His research interests include sleep pattern recognition, signal and image processing, neurofuzzy systems applied to the classification of physiological data, machine and deep learning. He was engaged in research on automated sleep-pattern detection and respiratory signal analysis, fitness for duty, human fatigue, drowsiness, alertness and performance.
\end{IEEEbiography} 

\begin{IEEEbiography}[{\includegraphics[width=1in,height=1.25in,clip,keepaspectratio]{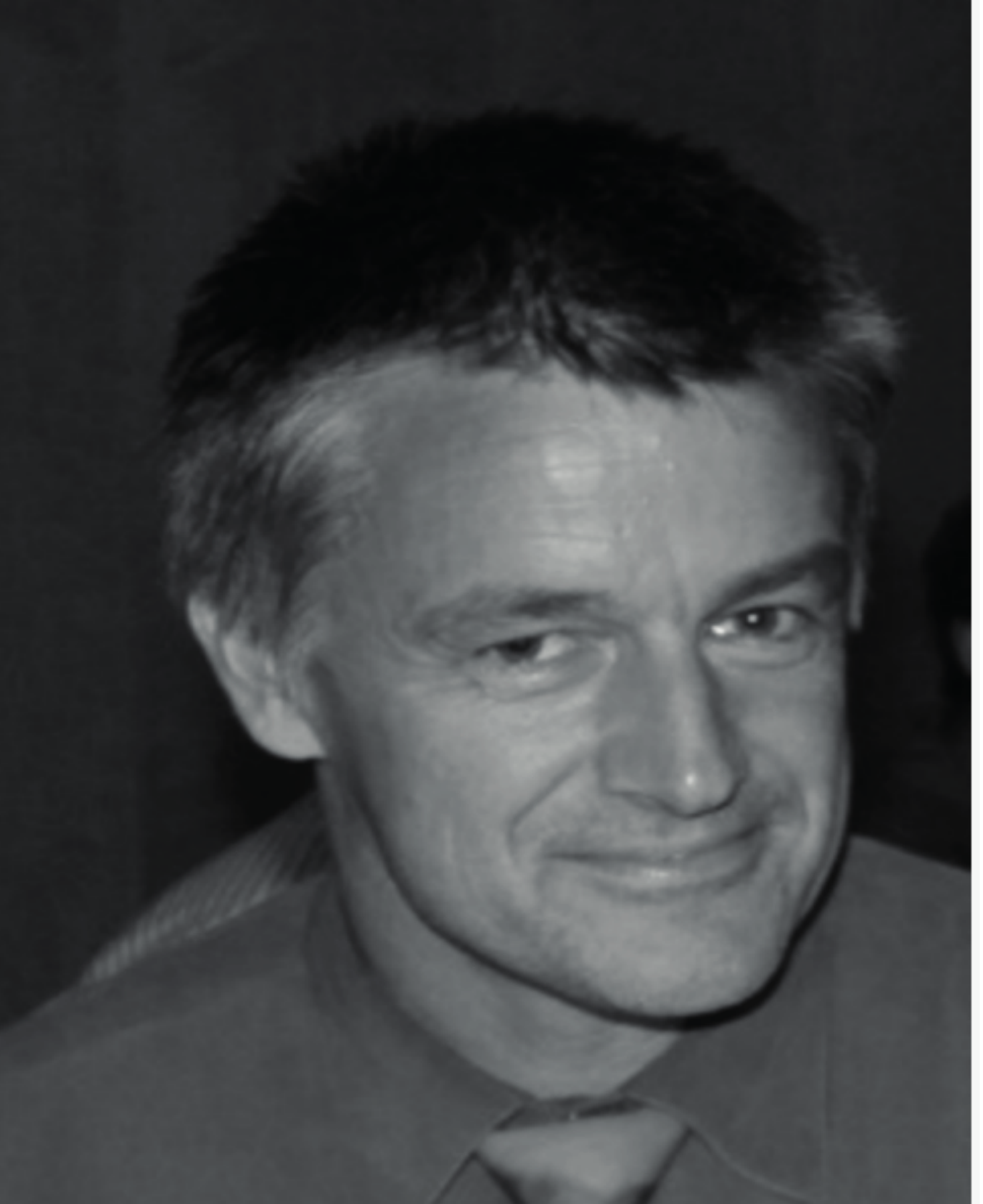}}]{Christoph Busch} is member of the Department of Information Security and Communication Technology (IIK) at the Norwegian University of Science and Technology (NTNU), Norway. He holds a joint appointment with the computer science faculty at Hochschule Darmstadt (HDA), Germany. Further he lectures the course Biometric Systems at Denmark’s DTU since 2007. On behalf of the German BSI he has been the coordinator for the project series BioIS, BioFace, BioFinger, BioKeyS Pilot-DB, KBEinweg and NFIQ2.0. In the European research program he was initiator of the Integrated Project 3D-Face, FIDELITY and iMARS. Further he was/is partner in the projects TURBINE, BEST Network, ORIGINS, INGRESS, PIDaaS, SOTAMD, RESPECT and TReSPAsS. He is also principal investigator in the German National Research Center for Applied Cybersecurity (ATHENE). Moreover Christoph Busch is co-founder and member of board of the European Association for Biometrics (www.eab.org) that was established in 2011 and assembles in the meantime more than 200 institutional members. Christoph co-authored more than 500 technical papers and has been a speaker at international conferences. He is member of the editorial board of the IET journal.
\end{IEEEbiography}

\end{document}